%% file: main.tex
\documentclass{bmvc2k}

\title{Emo-LiveSync: Real-Time Universal Lip Synchronization with Explicit Emotion Control}

\addauthor{Sangjoon Yu$^{\mathbf{*}}$}{sangjoon.yu@klleon.io}{1}
\addauthor{Seungeun Lee$^{\mathbf{*}}$}{seungeun.lee@klleon.io}{1}
\addauthor{JoonHyeok Shin}{shinjh0729@g.skku.edu}{2}
\addauthor{Ji-Su Kang}{jisu.kang@klleon.io}{1}

\addinstitution{
 Klleon AI Research \\
 Seoul, South Korea \\
}
\addinstitution{
 Sungkyunkwan University \\
 Suwon, South Korea \\[0.5em]
 $^{\mathbf{*}}$Equally contributed co-first authors.
}
\runninghead{Yu and Lee et al.}{Emo-LiveSync}

\usepackage{graphicx}
\usepackage{booktabs}

\usepackage[accsupp]{axessibility}

\usepackage{bbding}
\usepackage{makecell}
\usepackage{caption}
\usepackage{wrapfig}

\usepackage{hyperref}

\begin{document}

\maketitle

\begin{abstract}
Lip synchronization, the task of generating realistic lip movements synchronized with audio signals, is fundamental to applications such as film dubbing, virtual avatars, and digital content creation. Recent GAN-based methods offer computational efficiency but often lack visual fidelity, while diffusion-based approaches achieve high-quality results at the cost of substantial computational overhead. Moreover, most existing methods are limited to photo-realistic human faces, restricting their applicability to stylized characters and non-human entities. In this paper, we propose an efficient universal lip-sync video generation framework that achieves high-quality results without compromising computational efficiency. We build on LivePortrait, a warp-based portrait animation model, and introduce a training strategy that aligns audio signals to an implicit motion space by distilling knowledge from a pre-trained video-to-motion model. This enables real-time inference while maintaining high visual fidelity, and naturally generalizes to diverse stylistic characters beyond photo-realistic humans. Furthermore, we introduce an expressive control gate that enables explicit manipulation of emotional expressions for content enrichment. Extensive experiments demonstrate that our method achieves state-of-the-art performance in both visual quality and lip-sync accuracy while ensuring real-time inference.
\end{abstract}

\input{sec/01_intro}

\input{sec/02_rel}
\input{sec/03_method}
\input{sec/04_experiment}
\clearpage
\input{figs/10_qual_comparisons_sub}

\clearpage
\bibliography{main}
\end{document}

%% file: sec/01_intro.tex
\section{Introduction}
\input{figs/01_teaser}

Lip synchronization, the task of generating realistic lip movements synchronized with given audio signals, has emerged as a fundamental problem in computer vision and multimedia applications. This technology serves as a cornerstone for various downstream applications, including film dubbing, virtual avatars, video conferencing, and digital content creation. Despite its practical importance and growing demand, achieving high-quality lip-synced video generation remains a challenging problem due to the inherent complexity of modeling the intricate mapping between audio signals and corresponding facial movements.

Recent advances in generative models have led to significant progress in lip synchronization. GAN-based approaches \cite{prajwal2020lip,zhang2023dinet,zhong2023identity,guan2024resyncer} have demonstrated the ability to generate lip-synced videos with reasonable efficiency, yet they often struggle to achieve high visual fidelity, particularly in preserving fine-grained facial details and temporal consistency. On the other hand, diffusion-based methods \cite{zhong2024style,yu2024make,liu2024diffdub,li2024latentsync,ma2025sayanything,mukhopadhyay2024diff2lip,peng2025omnisync} have shown remarkable capability in producing high-fidelity results by leveraging the powerful generative capacity of large-scale diffusion models. However, these approaches typically entail substantial computational costs, making them impractical for real-time or resource-constrained applications. Consequently, an efficient approach that does not compromise the visual quality of generated lip-synced videos remains an open challenge.

Furthermore, most existing lip synchronization methods have been primarily developed and evaluated on photo-realistic human faces. This limitation significantly restricts their applicability to universal domains, such as stylized characters, animated avatars, or non-human entities. As the demand for lip synchronization in animation, gaming, and virtual content creation continues to grow, developing methods that generalize beyond photo-realistic humans becomes increasingly important.

To address these challenges, we revisit warp-based lip synchronization methods that offer an attractive balance between efficiency and visual quality. Specifically, we focus on learning a model that maps audio signals to an implicit motion space.
By leveraging LivePortrait~\cite{guo2024liveportrait}, a pre-trained video-to-motion model originally designed for portrait image animation,
we propose to align visual and audial features into a shared motion space represented by implicit keypoints. This approach enables efficient lip-synced video generation while maintaining high visual fidelity inherited from the warp-based transformation, and naturally generalizes to diverse character styles beyond photo-realistic humans.

In addition to achieving efficient and high-fidelity lip synchronization, we further extend our framework with an expressive control gate that enables explicit manipulation of emotional expressions. While most existing methods focus solely on synchronizing lip movements with audio, they overlook the potential for enriching generated content with controllable emotional cues. Our expressive control mechanism allows users to specify emotional transitions through a frame-wise blending schedule, facilitating diverse and expressive content generation for applications such as dubbing services and virtual character animation.
To comprehensively validate these capabilities across linguistically and visually diverse scenarios, we also present UniVox-Bench, a new evaluation benchmark that encompasses multilingual speech and diverse subjects including stylistic non-human characters, going beyond the English-centric and human-only settings of prior benchmarks.

Our contributions can be summarized as follows. First, we propose an efficient lip-sync video generation framework that aligns audio signals to an implicit motion space, achieving real-time inference without compromising visual fidelity. Second, we introduce a training strategy that distills knowledge from a pre-trained video-to-motion model, enabling robust audio-to-motion mapping for universal lip synchronization across diverse stylistic characters. Third, we present an expressive control gate that allows explicit manipulation of emotional expressions, thereby facilitating diverse and enriched content generation. Finally, we construct UniVox-Bench, a multilingual evaluation benchmark featuring diverse human and non-human stylistic subjects, which addresses the limited linguistic and visual coverage of existing lip synchronization benchmarks.

\vspace{-0.5cm}

%% file: figs/01_teaser.tex
\begin{figure}[t]
  \centering
  \includegraphics[width=0.99\linewidth]{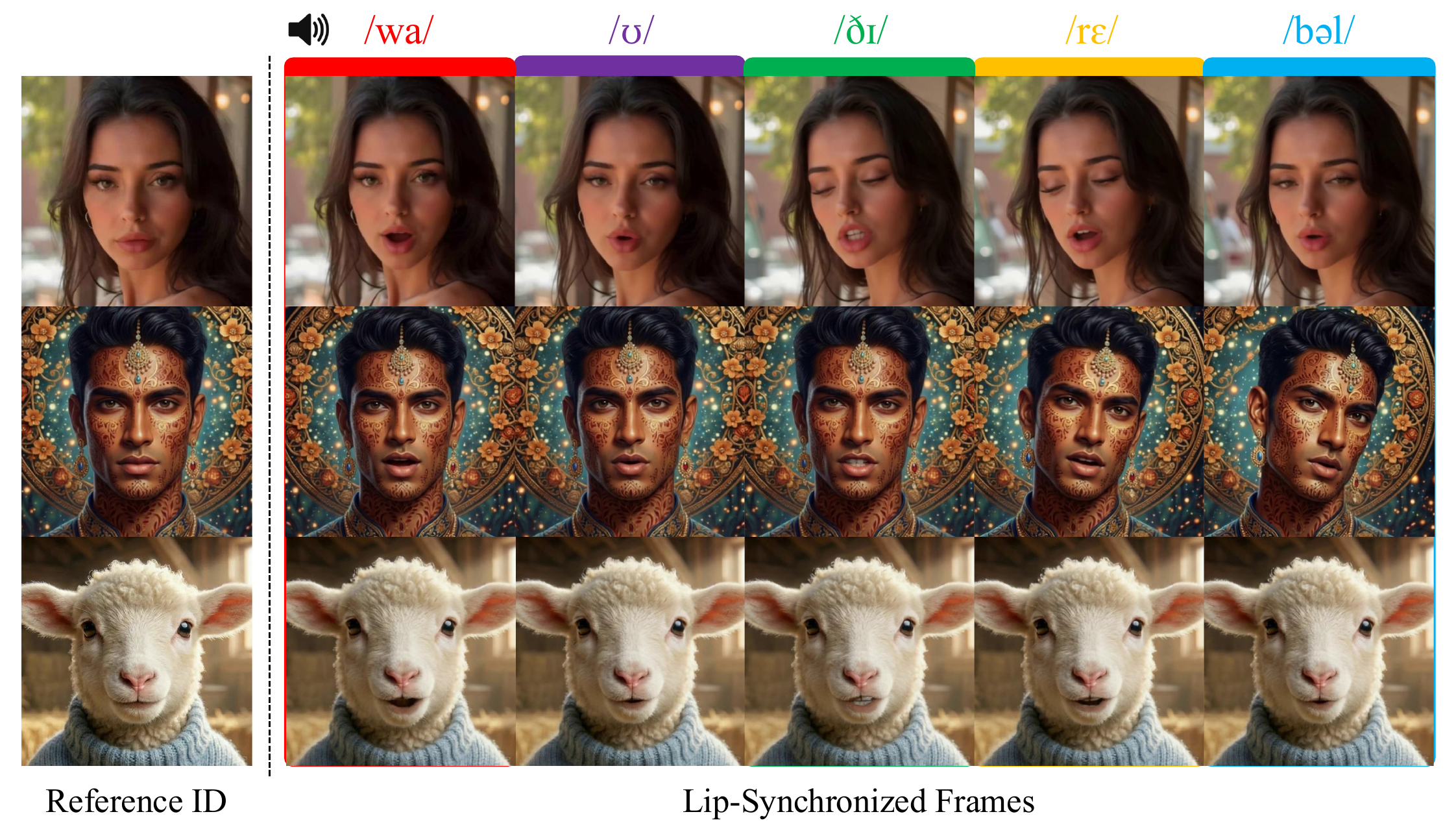}
  \vspace{0.4cm}
  \caption{We present an universal model that generates lip-synchronized video on stylistic humans and characters corresponding to driving audio sequences. Our method ensures computational efficiency without sacrificing high visual fidelity.}
  \vspace{-0.5cm}
\end{figure}

%% file: sec/02_rel.tex
\section{Related Work}

\input{figs/02_motivation}

\subsection{Audio-driven Talking Head Generation}
Audio-driven talking head generation has undergone several paradigm shifts.
Early methods~\cite{karras2017audio,jamaludin2019you,wiles2018x2face,zhou2020makelttalk} learned person-specific audio-to-face mappings via CNNs and RNNs.
Adversarial training~\cite{zhou2019talking,chen2019hierarchical,vougioukas2020realistic,zhou2021pose,lu2021live} then enabled one-shot synthesis for unseen identities.
Several approaches introduced explicit 3D priors~\cite{thies2020neural,zhang2021facial,peng2024synctalk,ye2024real3d,wei2024aniportrait,zhang2023sadtalker,tang2025real,sun2025vividtalk} or implicit 3D representations~\cite{guo2021ad,shen2022learning,cho2024gaussiantalker} with free-viewpoint rendering, while Transformers and 3DMM pipelines improved temporal coherence, and NeRF and 3DGS-based representation increased photorealism~\cite{moon2025geoavatar,lee2026secondary,lee2026audioavatar}.
Recent works such as \cite{shen2023difftalk,ma2023dreamtalk,tian2024emo,xu2024vasa,xu2024hallo,wang2024v,liu2024anitalker,cui2024hallo2,gao2025wan,jiang2024loopy,cui2025hallo3,chen2025echomimic,wang2025fantasytalking,tian2025emo2,chen2025hunyuanvideo,gan2025omniavatar,wei2025mocha,lin2025omnihuman} utilize diffusion-based architectures to synthesize realistic talking head videos from a single reference image and an audio stream.

However, talking head generation differs fundamentally from lip synchronization. The former lacks constraints to preserve source video integrity, often redefining the entire spatial-temporal composition. This is unsuitable for pipelines requiring fixed poses. In contrast, lip-sync operates within a video-to-video framework, requiring localized modifications that maintain consistency with original facial features and lighting.
As generative pipelines evolve, specialized lip synchronization becomes essential as a surgical tool to rectify phonetic-visual misalignments without altering the original motion dynamics or artistic intent.
Consequently, dedicated lip synchronization remains a distinct and essential research challenge.
\vspace{-0.5cm}
\subsection{Audio-driven Lip Synchronization}

Unlike audio-driven talking head generation that synthesizes full facial motion from a single image, lip synchronization is a video-to-video editing task that modifies only the mouth region of a source video to match target audio, preserving pose, expression, and background.
Early GAN-based methods~\cite{chung2016out,prajwal2019lipgan,song2019talking,vougioukas2018temporal} established encoder-decoder architectures with lip-sync discriminators.
Wav2Lip~\cite{prajwal2020lip} showed that a frozen pre-trained SyncNet discriminator is critical for sync accuracy, becoming the standard baseline.
Subsequent works improved fidelity via landmarks~\cite{zhong2023identity,wang2023talklip}, deformation fields~\cite{zhang2023dinet}, codebooks~\cite{wang2023lipformer}, style-based generators~\cite{guan2023stylesync,ki2023stylelipsync,guan2024resyncer}, and multi-stage pipelines~\cite{cheng2022videoretalking,kim2019neural,park2022synctalkface}.
Diff2Lip~\cite{mukhopadhyay2024diff2lip} first applied pixel-space diffusion to lip-sync inpainting, surpassing GANs in perceptual quality.
MuseTalk~\cite{zhang2024musetalk} achieves real-time inference via single-step latent prediction with Whisper features.
LatentSync~\cite{li2024latentsync} introduces end-to-end latent diffusion with StableSyncNet and temporal alignment to address shortcut learning.
OmniSync~\cite{peng2025omnisync} proposes mask-free Diffusion Transformers for generalization to stylized and AI-generated content.
Recent works~\cite{bigata2025keysync,zhang2025syncanyone,ma2025sayanything} further explore leakage-free generation and self-bootstrapping training.

However, existing methods generally face a trade-off among quality, efficiency, and generalization: diffusion-based approaches achieve high fidelity but incur substantial computational costs, while most methods remain limited to human faces and offer no control over emotions. In contrast, our approach aligns audio to an implicit motion space via knowledge distillation from a pre-trained video-to-motion model, enabling real-time inference that naturally generalizes to stylized characters and further supports explicit emotion control.
\vspace{-0.3cm}
\subsection{Warp-based Image Portrait Animation}

Unlike audio-driven lip synchronization focusing on mouth regions, portrait animation transfers full facial motion from a driving source to a target image; while diffusion-based methods~\cite{ma2024followyouremoji,xie2024xportrait,han2024faceadapter,wei2024aniportrait,hu2024animateanyone,xu2024magicanimate,cui2025hallo3,xu2025hunyuanportrait,li2025personalive} show promise, we focus on warping-based approaches.
Early works used 3DMMs or learned warps~\cite{thies2016face2face,averbuchelor2017bringing,kim2018deepvideo,wiles2018x2face,wu2018reenactgan,ren2021pirenderer,doukas2021headgan}.
Building on unsupervised keypoints~\cite{siarohin2018deformable,siarohin2019monkeynet}, FOMM~\cite{siarohin2019first} modeled motion via local affine transforms, extended with iterative refinement~\cite{nirkin2019fsgan,ha2020marionette}, region decomposition~\cite{siarohin2021mraa}, latent navigation~\cite{wang2022lia}, depth awareness~\cite{hong2022depth}, and thin-plate splines~\cite{zhao2022thin}.
Recent methods use 3D keypoints~\cite{wang2021facevid2vid} for volumetric warping at high resolution~\cite{drobyshev2022megaportraits,khakhulin2022rome,yin2022styleheat,zhang2023metaportrait,drobyshev2024emoportraits,guo2024liveportrait}.

These methods demonstrate distinct advantages in disentangling appearance from motion properties, thereby enabling the capture and control of high-fidelity facial expressions while rigorously preserving the identity of the reference image. In this work, we build upon a pre-trained warping-based portrait animation model by adapting it as video-to-motion model. We propose a knowledge distillation strategy that transfers the mapping of implicit keypoints from the driving video domain to the audio domain, facilitating real-time and high-fidelity lip-sync synthesis. Our method ensures robust, identity-consistent results across various stylistic characters and avatars without sacrificing computational efficiency.
\vspace{-0.3cm}
\subsection{Emotion Controllability}

Early works on emotional talking heads relied on categorical emotion labels~\cite{eskimez2022speech,wang2020mead,li2021writeaspeaker}, which limits fine-grained expressiveness. To address this, subsequent methods disentangled lip motion from upper-face expression for independent emotion manipulation~\cite{ji2021audio,ji2022eamm,peng2023emotalk,tan2024edtalk,gan2023eat}, with EMOTE~\cite{danecek2023emote} further introducing decoupled speech-emotion losses on FLAME to support both emotion type and intensity control. Controllable emotion has also been pursued in neighboring modalities. FaceCrafter~\cite{mishima2025facecrafter} inserts lightweight control modules into the cross-attention layers of an identity-conditional diffusion model, while IndexTTS2~\cite{zhou2026indextts2}, EmoDubber~\cite{cong2025emodubber}, and EmoSteer-TTS~\cite{xie2025emosteer} act on the speech side, respectively disentangling emotion from speaker timbre, exposing emotion type and intensity for movie dubbing, and converting emotion without any training. These works establish the criteria we adopt: semantic fidelity, continuous rather than discrete intensity, disentanglement from identity and content, and temporal consistency.

Our setting differs on two axes: these methods act on speech or on a static image, whereas we manipulate the visual emotion of an existing video whose head motion and background must be preserved; and they bind control to a specific trained backbone, through learned adaptors or its internal activations. In contrast, our module is training-free and model-agnostic: it synthesizes continuous per-frame expression displacements through parametric activation envelopes and gating variables, requiring neither learned parameters nor reference videos and enabling plug-and-play integration with existing lip-sync backbones.

%% file: figs/02_motivation.tex
\begin{figure}[t]
\centering
\begin{minipage}{0.46\textwidth}
    \small
    \centering

    \resizebox{\linewidth}{!}{%
    \setlength{\tabcolsep}{3pt}
    \begin{tabular}{@{}l|cc|c@{}}
      \toprule
      {Method} & \makecell{Real-time} & \makecell{Universal} & \makecell{Expressive} \\
      \midrule
      Wav2Lip \cite{prajwal2020lip}            & \CheckmarkBold & \XSolidBrush   & \XSolidBrush   \\[-0.5ex]
      Diff2Lip \cite{mukhopadhyay2024diff2lip} & \XSolidBrush   & \XSolidBrush   & \XSolidBrush   \\[-0.5ex]
      MuseTalk \cite{zhang2024musetalk}        & \CheckmarkBold & \XSolidBrush   & \XSolidBrush   \\[-0.5ex]
      LatentSync \cite{li2024latentsync}       & \XSolidBrush   & \XSolidBrush   & \XSolidBrush   \\[-0.5ex]
      OmniSync \cite{peng2025omnisync}         & \XSolidBrush   & \CheckmarkBold & \XSolidBrush   \\
      \midrule
      \textbf{Ours}                            & \CheckmarkBold & \CheckmarkBold & \CheckmarkBold \\
      \bottomrule
    \end{tabular}%
    }
    \vspace{0.4cm}
    \captionof{table}{Functional comparison across methods. Our approach supports both real-time inference and universal lip synchronization for diverse stylistic characters. Furthermore, our method enables explicit user control over expressions, thereby enriching the contents.}
\end{minipage}%
\hfill
\begin{minipage}[h]{0.50\textwidth}
    \centering
    \includegraphics[width=0.95\linewidth]{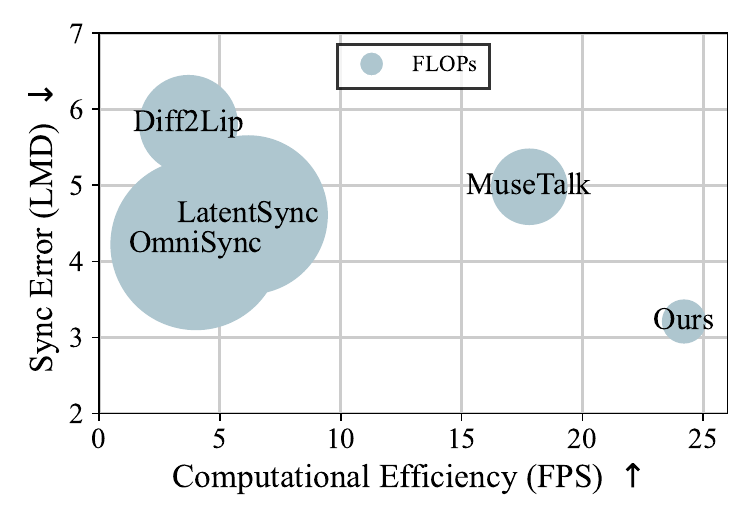}
    \caption{Comparison of lip-sync accuracy (LMD) versus computational time (FPS) against existing methods with model size. Our approach ensures lip-sync generation without compromising between visual high-fidelity and computational efficiency.}
\end{minipage}
\vspace{-0.5cm}
\end{figure}

%% file: sec/03_method.tex
\vspace{-0.3cm}
\section{Method}
\subsection{System Overview}

We present an efficient lip synchronization framework that generates high-fidelity talking videos from audio signals. Our approach builds upon a warp-based portrait animation method, LivePortrait~\cite{guo2024liveportrait}, which offers an attractive trade-off between computational efficiency and visual quality. The key insight is that audio and video signals for lip synchronization can be aligned in a shared motion space represented by implicit keypoints. It is illustrated in Fig. \ref{fig:alignment}.

Our framework consists of five main components: (1) an audio-to-motion model $\mathcal{M}_{A}$ that predicts implicit keypoints from audio signals, (2) a video-to-motion model $\mathcal{M}_{V}$ that extracts implicit keypoints from video frames, (3) a warp-based transformation module that computes dense motion fields, and (4) a video decoder that synthesizes the final output. To enable expressive content generation, we additionally introduce (5) an expressive control gate that allows explicit manipulation of emotional expressions.

During training, we align the audio-to-motion model to the implicit motion space defined by the pre-trained video-to-motion model~\cite{guo2024liveportrait}. Given paired audio-video data, the video-to-motion model extracts target keypoints $\mathbf{k}_{V}$ from ground-truth video frames, while the audio-to-motion model predicts keypoints $\mathbf{k}_{A}$ from the corresponding audio. The training objective minimizes the discrepancy between these two keypoint representations through a combination of Charbonnier loss for robust alignment, velocity loss for temporal coherence, and Wing loss for precise localization.

At inference time, given an arbitrary input video and a driving audio signal, the reference encoder first extracts appearance features from the input frames. The audio-to-motion model then predicts implicit keypoints synchronized with the audio, while the video-to-motion model provides source keypoints from the input video. The warp-based transformation module computes the motion field between source and predicted keypoints, which warps the reference features accordingly. Finally, the video decoder synthesizes lip-synced output frames that preserve the identity of the input while faithfully reflecting the audio-driven lip movements.
Fig. \ref{fig:overview} illustrates the architectural overview.

\input{figs/03_overview_and_alignment}

\subsection{Audio-Visual Alignment to Implicit Motion Space}

Our goal is to generate lip-synced videos synchronized with speaking audio signals while ensuring both computational efficiency and high visual fidelity. To achieve this, we depart from recent large-scale generative models based on diffusion processes that entail heavy computational costs, and instead revisit warp-based generative approaches.

We start from a key hypothesis: audio and video signals for lip synchronization meet in a shared motion space represented by implicit keypoints. Building upon this hypothesis, we align an audio-to-motion model $\mathcal{M}_{A}$ to the motion space defined by implicit keypoints $\mathbf{k}$ extracted from a pre-trained video-to-motion model $\mathcal{M}_{V}$ \cite{guo2024liveportrait}
, as illustrated in Fig. \ref{fig:alignment}.
This is formulated as minimizing the discrepancy between $\mathbf{k}_{A}$ predicted by $\mathcal{M}_{A}$ and $\mathbf{k}_{V}$ predicted by $\mathcal{M}_{V}$. The basic alignment objective is defined as the L2 loss: $\left\| \mathbf{k}_{A} - \mathbf{k}_{V} \right\|_{2}^{2}$,
where $\mathbf{k}_{A} = \mathcal{M}_{A}(\mathbf{a})$ denotes the implicit keypoints predicted from audio input $\mathbf{a}$, and $\mathbf{k}_{V} = \mathcal{M}_{V}(\mathbf{v})$ denotes those extracted from the corresponding video frames $\mathbf{v}$.

To facilitate stable training when aligning two high-dimensional raw signals from different modalities, we extend the objective to the Charbonnier loss \cite{barron2019general} formulation:
\begin{equation}
    \mathcal{L}_{kpt} = \sqrt{\left\| \mathbf{k}_{A} - \mathbf{k}_{V} \right\|^{2} + \epsilon^{2}},
\end{equation}
where $\epsilon$ is a small constant for numerical stability.
The variant of the L1 loss
provides robustness against outliers while maintaining smooth gradients near zero, which proves beneficial when bridging the heterogeneous distributions of audio and visual modalities.

To encourage temporally coherent motion and reduce jittering artifacts, we introduce a velocity loss that penalizes discrepancies in keypoint dynamics:
\begin{equation}
    \mathcal{L}_{vel} = \| ({\mathbf{k}}^t_V - {\mathbf{k}}^{t-1}_V) - (\mathbf{k}^t_A - \mathbf{k}^{t-1}_A) \|_2^2
\end{equation}
This loss ensures that the temporal changes in predicted keypoints from audio signals match those from visual signals, promoting smooth transitions between consecutive frames.

We additionally employ the Wing loss~\cite{feng2018wing}, which has proven effective for landmark localization. It is defined as:
\begin{equation}
    \mathcal{L}_{wing} = \begin{cases}
        w \ln(1 + |x|/\kappa) & \text{if } |x| < w \\
        |x| - C & \text{otherwise}
    \end{cases}
\end{equation}
where $x = {\mathbf{k}}^t_V - \mathbf{k}^t_A$, $w$ and $\kappa$ are hyperparameters controlling the nonlinear region, and $C = w - w\ln(1 + w/\kappa)$ ensures continuity. The Wing loss applies stronger gradients to small errors compared to L2 loss, enabling more precise keypoint localization. The final training objective combines all three losses:
\begin{equation}
    \mathcal{L}_{total} = \lambda_{kpt}\mathcal{L}_{kpt} + \lambda_{vel}\mathcal{L}_{vel} + \lambda_{wing}\mathcal{L}_{wing}
\end{equation}
where $\lambda_{kpt}$, $\lambda_{vel}$, and $\lambda_{wing}$ are weighting factors balancing the contribution of each term.\\
\vspace{-0.5cm}
\paragraph{Architecture.}
\input{figs/04_arc_a2m}
Our audio-to-motion model $\mathcal{M}_{A}$ is designed to align audio sequences to the shared implicit motion space. We build our framework upon a pre-trained warp-based portrait animation model, borrowing its transformation module and video decoder, and focus our contribution on bridging the audio modality to the motion space.
As illustrated in Fig.~\ref{fig:arch_audio2motion}, $\mathcal{M}_{A}$ adopts a hybrid Transformer-LSTM architecture. Given a raw audio, we first extract phonetic and prosodic representations using a pre-trained Wav2Vec~\cite{baevski2020wav2vec}. The features are then processed by a stack of Transformer encoder blocks to capture long-range contextual dependencies, followed by an LSTM layer that models the sequential dynamics of speech for smooth and coherent motion generation. Finally, the LSTM outputs are aggregated via attention pooling and projected through a linear layer to produce implicit keypoints $\mathbf{k}_{A} \in \mathbb{R}^{N \times D}$, where $N$ and $D$ denote the number and dimension of keypoints, respectively. It is noteworthy that the full set of $N = 21$ implicit keypoints defined in \cite{guo2024liveportrait} is retained rather than restricting the prediction to a mouth-region subset, because lip articulation is not independent of the rest of the face: jaw rotation, cheek deformation, and the surrounding facial geometry co-vary with mouth motion, and predicting the complete keypoint set allows the model to represent these dependencies instead of severing them.

\subsection{Expressive Control Gate}
Audio-driven lip synchronization produces plausible mouth shapes from speech, yet the resulting faces often appear emotionally neutral because speech audio primarily encodes phonemic content; affective cues such as brow raises or cheek deformations are only weakly correlated with the acoustic signal.
Prior methods~\cite{gan2023eat,danecek2023emote,ji2022eamm,tan2024edtalk} address this by training dedicated emotion encoders tied to specific architectures.
We instead propose a \emph{training-free, model-agnostic} emotion injection module based on signal processing, requiring no learned parameters or reference video.

We define two latent control signals: an activation envelope $u^t \!\in\! [0,1]$ controlling expression intensity, and a gating variable $v^t \!\in\! [0,1]$ determining the distribution between two expression states $\boldsymbol{\delta}_1, \boldsymbol{\delta}_2 \!\in\! \mathbb{R}^{N \times D}$. The final blending weights are $w_1^t = u^t(1 - v^t)$ and $w_2^t = u^t \cdot v^t$, yielding the emotion-injected keypoints:
\begin{equation}
    \hat{\mathbf{k}}_A^t = \mathbf{k}_A^t + w_1^t\,\boldsymbol{\delta}_1 + w_2^t\,\boldsymbol{\delta}_2,
\end{equation}
which are passed to the warp-based transformation module. The two states $\boldsymbol{\delta}_1, \boldsymbol{\delta}_2$ are fixed offsets constructed once, offline, and are never derived from the video being edited: for each target emotion we extract implicit keypoints with the frozen $\mathcal{M}_V$ from training-set frames (Sec.~\ref{sec:setup}) and take the difference between an expressive and a neutral frame of the same subject, so that identity-specific geometry cancels and only the expression component remains. Because the conveyed emotion is a perceptual rather than a geometric attribute, the representative pair is determined by human verification instead of an automatic recognizer, avoiding the domain bias of a classifier trained on photo-realistic faces; annotators confirm that a candidate is perceived as the intended emotion and that its displacement magnitude is neither imperceptible nor exaggerated. The selected states are then frozen and blended identically into every identity's audio keypoints, and are added on top of $\mathbf{k}_A^t$ rather than replacing it, so the gate carries no learned parameters, preserves the phoneme-driven lip motion, and admits no test-video content into the model.

\input{figs/ecg}

The base activation envelope $u_{\mathrm{base}}$ is constructed by concatenating rise--hold--fall segments with cosine easing:
\begin{equation}
    u_{\mathrm{base}}(t) \!=\!
    \begin{cases}
        \tfrac{1}{2}\bigl(1 - \cos\bigl(\pi\tfrac{t - t_r}{t_h - t_r}\bigr)\bigr), & t_r \!\le\! t \!<\! t_h \\[2pt]
        1, & t_h \!\le\! t \!<\! t_f \\[2pt]
        \tfrac{1}{2}\bigl(1 + \cos\bigl(\pi\tfrac{t - t_f}{t_e - t_f}\bigr)\bigr), & t_f \!\le\! t \!<\! t_e \\[2pt]
        0, & \text{otherwise}
    \end{cases}
\end{equation}
where $t_r, t_h, t_f, t_e$ denote the rise start, hold start, fall start, and fall end times.
To enhance realism, we inject band-limited jitter with amplitude proportional to the activation level: $u_{\mathrm{jitter}}^t = u_{\mathrm{base}}^t\,(1 + \alpha \, n_{\mathrm{BL}}^t)$, where $\alpha$ controls jitter intensity and $n_{\mathrm{BL}}^t = \mathrm{BandPass}_{[f_l, f_h]}(\epsilon^t)$, $\epsilon^t \!\sim\! \mathcal{N}(0,1)$.
The signal is then stabilized via exponential moving average (EMA) smoothing followed by slew-rate limiting:
\begin{equation}
    \tilde{u}^t = \beta\,\tilde{u}^{t-1} + (1-\beta)\,u_{\mathrm{jitter}}^t, \quad
    u^t = \tilde{u}^{t-1} + \mathrm{clamp}\!\left(\tilde{u}^t - \tilde{u}^{t-1},\, -s_{\max},\, s_{\max}\right),
\end{equation}
where $\beta \!\in\! (0,1)$ is the smoothing factor and $s_{\max}$ bounds the maximum per-frame change, suppressing abrupt transitions.
The gating signal $v^t$ transitions smoothly between two expression states via cosine interpolation: $v^t = \tfrac{1}{2}\bigl(1 - \cos\bigl(\pi(t - t_s)/(t_e - t_s)\bigr)\bigr)$ for $t_s \!\le\! t \!\le\! t_e$, where $t_s$ and $t_e$ are the transition boundaries.
Finally, both $u$ and $v$ are resampled to match the target frame count $T$ via linear interpolation for $t \!=\! 0,\ldots,T\!-\!1$, enabling length-independent application across varying video durations.

%% file: figs/03_overview_and_alignment.tex
\begin{figure}[t]
\centering
\begin{minipage}[t]{0.48\textwidth}
    \centering
    \includegraphics[width=\linewidth]{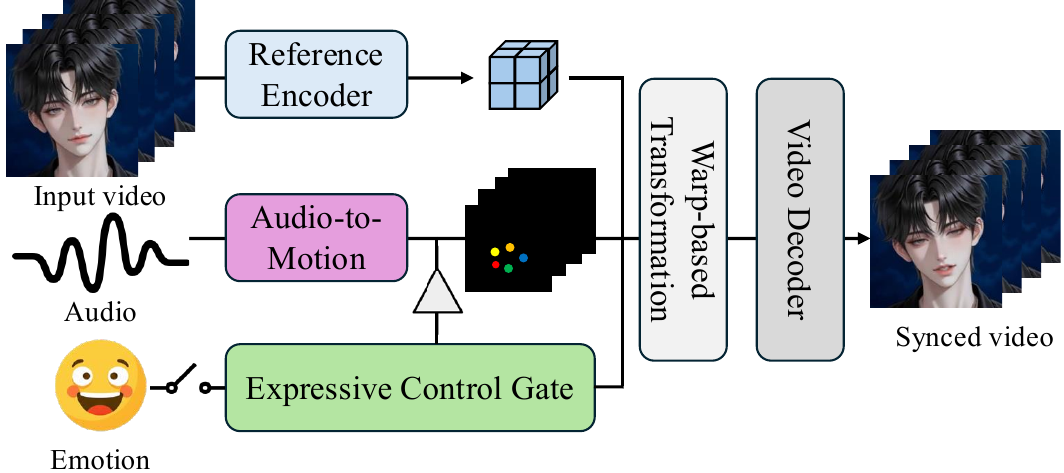}
    \vspace{0.4cm}
    \caption{System overview of our proposed lip-synced video generation framework at inference stage. Given a source video and driving audio, our model produces temporally coherent lip-synced video.}
    \label{fig:overview}
\end{minipage}%
\hfill
\begin{minipage}[t]{0.48\textwidth}
    \centering
    \includegraphics[width=\linewidth]{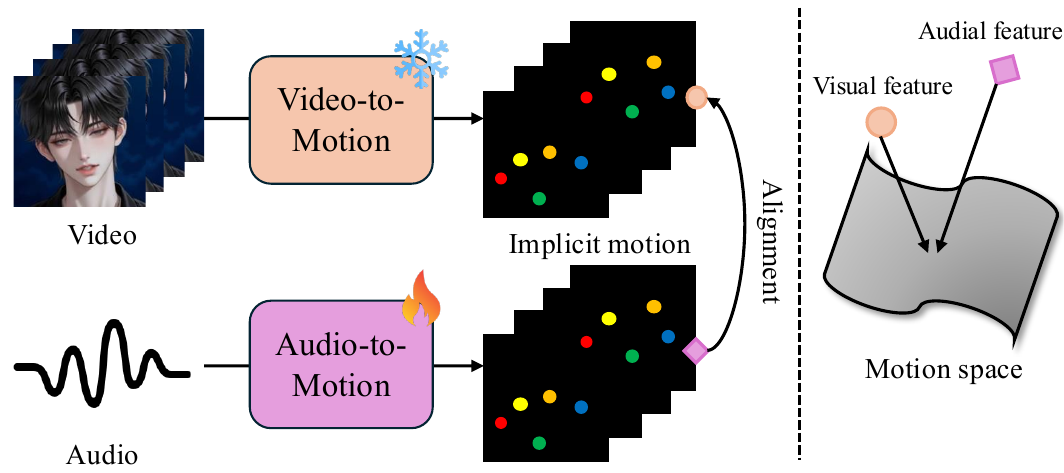}
    \vspace{0.4cm}
    \caption{Key concept of learning audio-to-motion model by aligning with video-to-motion model into an implicit motion space.}
    \label{fig:alignment}
\end{minipage}
\vspace{-0.3cm}
\end{figure}

%% file: figs/04_arc_a2m.tex
\begin{wrapfigure}[12]{r}[0pt]{0.5\textwidth}
    \vspace{-2.0em}
    \centering
    \includegraphics[width=0.48\textwidth]{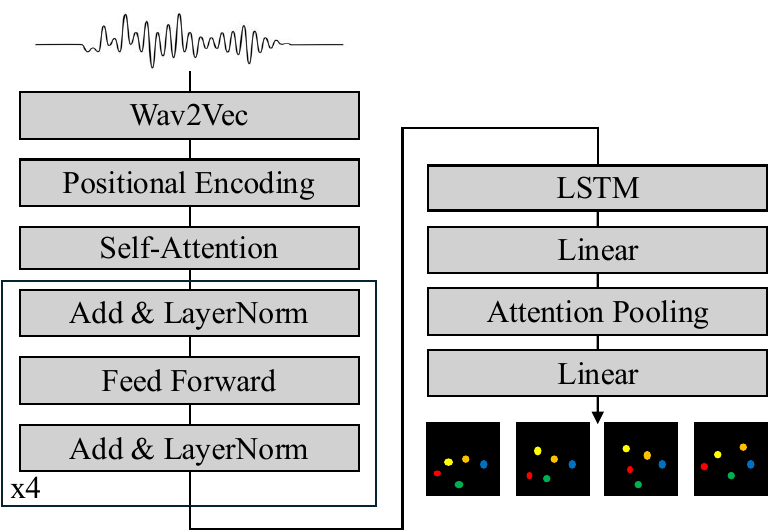}
    \vspace{0.3cm}
    \caption{Architecture flow of our proposed audio-to-implicit motion generation model.}
    \label{fig:arch_audio2motion}
\end{wrapfigure}

%% file: figs/ecg.tex
\begin{figure}[t]
  \centering
  \includegraphics[width=0.8\linewidth]{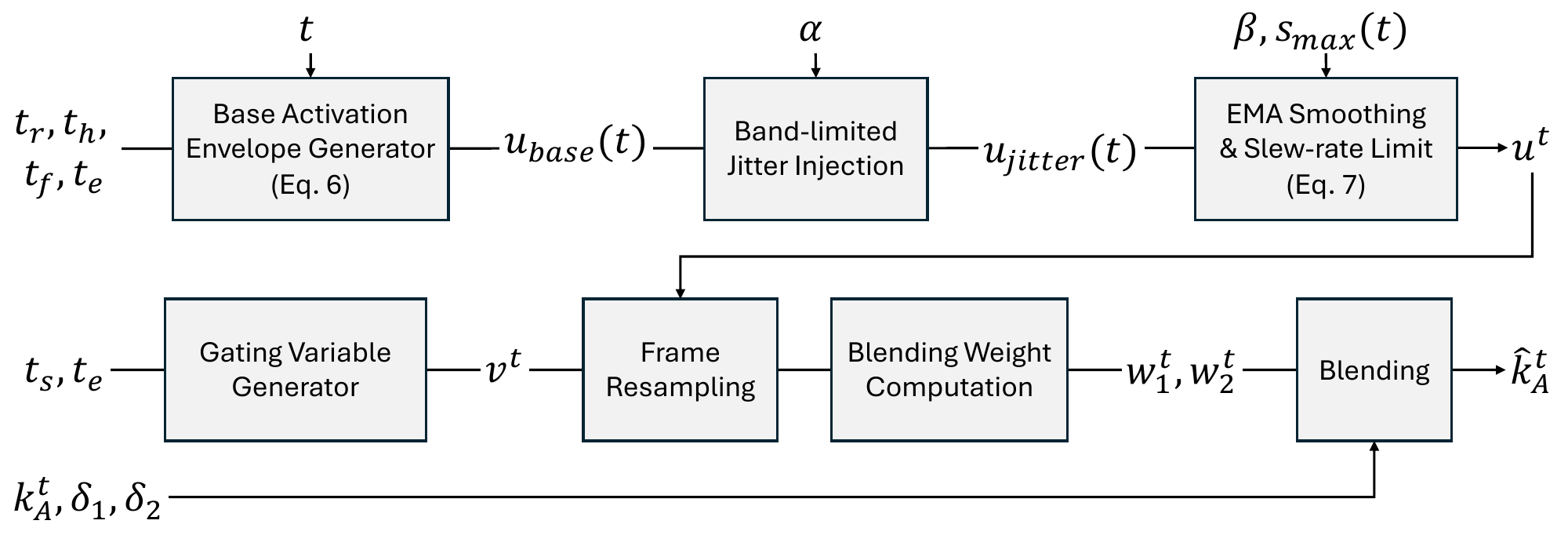}
  \vspace{0.5cm}
  \caption{Flow diagram of our proposed expressive control gate for injection and transition of emotion states.}
  \vspace{-0.5cm}
\end{figure}

%% file: sec/04_experiment.tex
\vspace{-0.5cm}
\section{Experiments}
\vspace{-0.2cm}
\subsection{Experimental Setup}
\label{sec:setup}
\paragraph{Dataset.}
We evaluate on two standard benchmarks: HDTF~\cite{zhang2021flow}, comprising 720p+ talking-face videos, and VoxCeleb2~\cite{chung2018voxceleb2}, comprising in-the-wild celebrity utterances. We randomly select 50 subjects in total, 20 from HDTF and 30 from VoxCeleb2, ensuring diversity in identity, pose, and audio characteristics. We further introduce \textit{UniVox-Bench}, since prior benchmarks~\cite{zhang2021flow,chung2018voxceleb2,peng2025omnisync} are restricted to predominantly English speech and/or real human subjects. It comprises 34 subjects, 46\% real humans and 54\% non-human stylistic characters including cartoon, anime and animals, paired with speech spanning seven languages: English, Chinese, French, Italian, Japanese, Korean, and Spanish. It is collected via web crawling and AI content generator with ElevenLabs~\cite{elevenlabs2024} and Gemini~\cite{geminiteam2024gemini}.
\vspace{-0.5cm}
\paragraph{Metrics.}
We evaluate visual fidelity, identity preservation, and lip synchronization. For visual fidelity, we report PSNR and LPIPS for per-frame reconstruction, FID~\cite{heusel2017gans} for distributional similarity to real frames, FVD~\cite{unterthiner2018towards} for temporal coherence, and HyperIQA~\cite{Su_2020_CVPR}, a no-reference metric that scores perceptual quality without ground truth. Identity preservation is measured by CSIM, the cosine similarity between identity embeddings of the source and generated videos. For lip synchronization, we use LMD, the landmark distance capturing geometric alignment of lip movements, and AVS~\cite{yaman2024eccv,yaman2024cvprw}, a reference-free score computing the cosine similarity between audio and visual features of a pretrained AV model~\cite{shi2022learning}. We exclude the SyncNet-based metric LSE-C~\cite{prajwal2020lip} on HDTF and VoxCeleb2 since several baselines train with SyncNet loss, which biases the score in their favor on these datasets.
\vspace{-0.5cm}
\paragraph{Implementation details.}
Our model takes as input a source video of arbitrary length with 512$\times$512 resolution, along with driving audio sampled at 16\,kHz. The audio-to-motion is aligned onto implicit motion space, which consists of $N = 21$ keypoints. For training, we utilize Lip Reading in the Wild (LRW) dataset \cite{chung2016lip} without additional data augmentation. The expression states $\boldsymbol{\delta}_1, \boldsymbol{\delta}_2$ of the expressive control gate are likewise built from the LRW training split, which is disjoint from evaluations; the offsets are thus constructed entirely outside the evaluation data and remain fixed across all subjects and benchmarks. The training objective is weighted with $\lambda_{kpt} = \lambda_{vel} = 1$ and $\lambda_{wing} = 1.5$, with the Wing loss parameters set to $w = 10$. We employ Adam optimizer \cite{kingma2014adam} with $\beta_1$ = 0.9 and $\beta_2$ = 0.99, and learning rate of 1e-4. The training is conducted on a single NVIDIA A100 GPU.

\input{tab/05_quan_hdtf_and_voxceleb2}

\input{tab/06_cr_and_quan_univox}
\vspace{-0.3cm}
\subsection{Comparison to State-of-the-Art Methods}

For fair comparison, the expressive control gate is disabled throughout the quantitative evaluation, so that Table~\ref{tab:main_comparison} \&~\ref{tab:newbench} measure lip synchronization alone under the same condition as the baselines. We compare our method with existing state-of-the-art lip-sync video generation methods \cite{prajwal2020lip,mukhopadhyay2024diff2lip,zhang2024musetalk,li2024latentsync,peng2025omnisync} on the HDTF and VoxCeleb2 datasets. For each method, we conduct evaluation using its publicly available code and checkpoints. For OmniSync, since only a web demo\footnote{\url{https://klingai.com/app/ai-human/video}} is available, we generate lip-synced videos through its online interface. As shown in Table~\ref{tab:main_comparison}, our method achieves the best performance across all metrics on both HDTF and VoxCeleb2 datasets. In terms of generation quality, we obtain the lowest FID and FVD scores, indicating superior frame-level realism and temporal coherence compared to all baselines. For reconstruction quality, our method attains the highest PSNR and CSIM while achieving the lowest LPIPS, demonstrating faithful preservation of visual details and source identity. Regarding lip synchronization, we achieve the lowest LMD and the highest AVS on both datasets, confirming that our method produces more accurate and temporally aligned lip movements.
In Table~\ref{tab:newbench}, we further evaluate on UniVox-Bench, a more challenging benchmark comprising multi-lingual speeches and stylistically diverse characters. Our method consistently outperforms all competitors across every metric, achieving the best FID, FVD, and HyperIQA scores for visual quality, as well as the highest LSE-C and AVS for lip-sync accuracy. These results demonstrate that our approach generalizes robustly beyond standard English-centric benchmarks to diverse linguistic and visual conditions.
In Table~\ref{tab:fps}, we report frames per second (FPS) and number of model parameters (\#Params) to evaluate computational efficiency. It is measured on a single RTX 3090 GPU. Since OmniSync does not provide officially available code, we cannot report its \#Params, and instead report FPS calculated from the time required to generate lip-synced videos through its available demo page. These results demonstrate that our method is computationally efficient, enabling real-time inference without compromising high visual fidelity.

\input{figs/07_qual_samples_and_multi-lingual}

\input{figs/08_qual_expression_and_userstudy}

\vspace{-0.3cm}
\subsection{Visual Analysis}
Fig.~\ref{fig:qual_samples} presents temporal lip-sync results across diverse human and non-human characters, demonstrating highly consistent universal synchronization. In Fig.~\ref{fig:comparison_human} and Fig.~\ref{fig:comparison_char}, we compare with GAN-based and diffusion-based methods; our approach achieves competitive visual quality while maintaining robust performance even for stylistic characters where most existing methods fail. These results suggest that aligning auditory and visual signals in an implicit motion space offers a compelling and efficient alternative to diffusion-based approaches. We further report quantitative multilingual comparisons on UniVox-Bench in Fig.~\ref{fig:multi-lingual1}, where existing methods exhibit acceptable synchronization only for limited languages or human subjects, while ours consistently achieves robust performance across all languages, validating the generalizability of our disentangled motion representation.
\vspace{-0.3cm}
\subsection{Ablation Study and Analysis}
In Fig.~\ref{fig:abl_loss}, we present ablation experiments on the loss functions designed for distilling knowledge from the video-to-implicit motion space mapping model to the audio-to-implicit motion space mapping model. We start from a baseline using L2 loss between implicit keypoints mapped from video and those mapped from audio. We then sequentially add $\mathcal{L}_{vel}$ and $\mathcal{L}_{wing}$. We evaluate on five subjects including stylistic characters, measuring PSNR for visual fidelity, CSIM for identity preservation, and LMD for lip synchronization. Compared to the baseline, the velocity loss improves performance by 1.1\% across all metrics. Furthermore, the wing loss yields an additional improvement of 8.7\% across all metrics. These results demonstrate the effectiveness of our proposed objectives for audio-to-motion mapping. We also investigate architecture design choices for the audio-to-implicit motion generation model in Fig.~\ref{fig:abl_arch}. We compare our proposed architecture against a Transformer-based variant and a plain MLP, evaluating convergence behavior across Charbonnier, velocity, and MSE losses. As shown, our design achieves both faster convergence and lower final loss values, exhibiting superior stability and optimality over the alternative configurations.
Beyond generating video content aligned with the context conveyed in the audio, we introduce an additional intriguing capability. Through an expressive control gate, we incorporate an optional emotion amplification component for diverse and rich content generation. In Fig.~\ref{fig:expression}, we demonstrate lip-synchronized videos generated from the same audio signal with three distinct emotions injected. The results show that our method produces expressive videos corresponding to the injected emotions.
To confirm that this expressiveness does not come at the cost of synchronization, Fig.~\ref{fig:abl_ecg} compares generated frames with and without the expressive control gate at matched phonemes. At each sampled phoneme, the mouth remains articulated consistently with the underlying speech in both settings, while the brows, cheeks, and eyes carry the injected emotion. Since the gate applies its offsets additively on top of the audio-predicted keypoints, the phoneme-driven articulation established by our alignment objective is retained, and emotion enters as a complementary rather than a competing signal. We anticipate that this proposed approach will serve as a compelling element for application to video dubbing services.

\input{figs/09_abl_loss_and_arch}
\input{figs/abl_ecg}
\vspace{-0.3cm}
\subsection{User Study}

We conduct an A/B test comparing our method against a state-of-the-art method~\cite{peng2025omnisync}. 19 participants rate their preference on a 5-point scale along three aspects: identity preservation, lip synchronization, and visual quality. As reported in Fig.~\ref{fig:user_study}, our method is preferred across all three, and a paired $t$-test on the preference scores confirms that the margin is significant ($p = 1.42 \times 10^{-9}$). Together with the efficiency reported in Tab.~\ref{tab:fps}, this validates the core motivation of this work, high visual fidelity without compromising computational efficiency, under human judgment rather than automatic metrics alone.
\input{figs/userstudy_emotion}
We further evaluate the expressive control gate with the same participants. Since automatic emotion metrics collapse continuous expression into discrete categories and overlook temporal naturalness, a central design objective of our gate, we assess it perceptually. For seven source identities spanning diverse ethnicities and genders, we inject happy, angry, and sad emotions, yielding 21 comparison pairs per participant, each rated on a 5-point mean opinion score scale with and without the gate. As reported in Fig.~\ref{fig:emotion_userstudy}, the gate raises the score from 1.87 to 4.03 for happy, 2.17 to 3.83 for angry, and 1.80 to 4.20 for sad, with an overall mean of 4.02 and lip synchronization reported as unaffected. Participants noted that the upturned mouth corners induced by larger lip openings occasionally weakened the perceived intensity of angry expressions, consistent with its lower score.

\vspace{-0.3cm}
\section{Conclusion}

In this paper, we propose an efficient universal lip-synced video generation method that achieves high-quality lip synchronization without compromising visual fidelity. Leveraging LivePortrait with pre-trained implicit keypoints, we introduce a training strategy that distills knowledge from a pre-trained video-to-motion model into audio-to-implicit-motion mapping, optimized with complementary Charbonnier, velocity, and Wing losses for stable cross-modal alignment. The consistent deformation provided by the warping module enables both high-fidelity results and real-time inference, while naturally generalizing to stylistic non-human characters beyond humans. We further propose a training-free, model-agnostic expressive control gate that explicitly manipulates emotions through parametric activation envelopes, enriching contents without additional supervision. Extensive experiments demonstrate that our approach is both robust and effective for lip synchronization, offering a compelling alternative to computationally expensive diffusion-based methods.

%% file: tab/05_quan_hdtf_and_voxceleb2.tex
\begin{table*}[t!]
\centering
\small
\caption{Quantitative comparison with existing lip-sync video generation methods on HDTF and VoxCeleb2 datasets. Best: \textbf{bolded}, and second: \underline{underlined}.}
\vspace{0.4cm}
\label{tab:main_comparison}
\resizebox{\textwidth}{!}{%
\setlength{\tabcolsep}{4pt}
\renewcommand{\arraystretch}{1.05}
\begin{tabular}{@{}l|ccccccc|ccccccc@{}}
\toprule
& \multicolumn{7}{c|}{HDTF} & \multicolumn{7}{c}{VoxCeleb2} \\
\cmidrule(lr){2-8} \cmidrule(lr){9-15}
& \multicolumn{2}{c}{Gen. Quality} & \multicolumn{3}{c}{Recon. Quality} & \multicolumn{2}{c|}{Lip-Sync} & \multicolumn{2}{c}{Gen. Quality} & \multicolumn{3}{c}{Recon. Quality} & \multicolumn{2}{c}{Lip-Sync} \\
\cmidrule(lr){2-3} \cmidrule(lr){4-6} \cmidrule(lr){7-8} \cmidrule(lr){9-10} \cmidrule(lr){11-13} \cmidrule(lr){14-15}
Methods
& FID$\downarrow$ & FVD$\downarrow$
& PSNR$\uparrow$ & LPIPS$\downarrow$ & CSIM$\uparrow$
& LMD$\downarrow$ & AVS$\uparrow$
& FID$\downarrow$ & FVD$\downarrow$
& PSNR$\uparrow$ & LPIPS$\downarrow$ & CSIM$\uparrow$
& LMD$\downarrow$ & AVS$\uparrow$ \\
\midrule
Wav2Lip \cite{prajwal2020lip}
& 12.50 & 304.35 & 30.65 & 0.135 & 0.818 & 7.88 & 0.3742
& 10.80 & 257.85 & 30.48 & 0.146 & 0.791 & 7.91 & 0.3658 \\
Diff2Lip \cite{mukhopadhyay2024diff2lip}
& 10.30 & 260.45 & 33.04 & 0.122 & 0.905 & 5.32 & 0.3886
& 9.80 & 210.45 & 34.82 & 0.057 & 0.932 & 5.80 & 0.3950 \\
MuseTalk \cite{zhang2024musetalk}
& 9.35 & 246.75 & 35.06 & 0.062 & 0.932 & 4.98 & 0.4103
& 7.10 & 203.43 & 34.57 & 0.064 & 0.905 & 4.98 & 0.4021 \\
LatentSync \cite{li2024latentsync}
& \underline{7.22} & \underline{151.00} & {35.16} & {0.039} & {0.958} & {3.86} & {0.5174}
& \underline{5.70} & {85.75} & {34.82} & \underline{0.042} & \underline{0.958} & {4.61} & {0.5081} \\
OmniSync \cite{peng2025omnisync}
& 13.75 & 169.44 & \underline{36.01} & \underline{0.032} & \underline{0.966} & \underline{3.17} & \underline{0.5352}
& 16.64 & \underline{71.38} & \underline{35.16} & {0.049} & {0.940} & \underline{4.22} & \underline{0.5288} \\
\midrule
\textbf{Ours}
& \textbf{5.64} & \textbf{66.77} & \textbf{36.67} & \textbf{0.026} & \textbf{0.983} & \textbf{2.74} & \textbf{0.5618}
& \textbf{4.64} & \textbf{52.26} & \textbf{36.41} & \textbf{0.021} & \textbf{0.985} & \textbf{3.21} & \textbf{0.5525} \\
\bottomrule
\end{tabular}%
}
\end{table*}

%% file: tab/06_cr_and_quan_univox.tex
\begin{table}[t!]
\centering
\vspace{-0.3cm}
\begin{minipage}[t]{0.48\textwidth}
    \small
    \centering
    \caption{Report of computational cost compared to the existing methods. It is measured on a single RTX 3090 GPU device. Best: \textbf{bolded}, and second: \underline{underlined}.}
    \vspace{0.4cm}
    \label{tab:fps}
    \resizebox{\linewidth}{!}{%
    \setlength{\tabcolsep}{3pt}
    \begin{tabular}{@{}l|ccccc@{}}
      \toprule
      Methods & LMD $\downarrow$ & FPS $\uparrow$ & Latency (ms) $\downarrow$ & \#Params $\downarrow$ & FLOPs $\downarrow$ \\
      \midrule
      MuseTalk \cite{zhang2024musetalk}   & 4.98 & \underline{11.71} & \underline{85.34} & 933.60M & \underline{1,330.92G} \\
      Diff2Lip \cite{mukhopadhyay2024diff2lip}   & 5.80 & 3.70 & 267.17 & \textbf{101.67M} & 3810.36G \\
      LatentSync \cite{li2024latentsync}  & 4.61 & 6.17 & 162.07 & 1,350.75M & 781.03T \\
      OmniSync \cite{peng2025omnisync}    & \underline{4.22} & 4.01 & 249.37 & - & - \\
      \textbf{Ours}                       & \textbf{3.21} & \textbf{24.00} & \textbf{41.66} & \underline{126.59M} & \textbf{649.71G} \\
      \bottomrule
    \end{tabular}%
    }
\end{minipage}%
\hfill
\begin{minipage}[t]{0.48\textwidth}
    \small
    \centering
    \caption{Quantitative comparison of visual quality and audio-visual synchronization metrics on UniVox including multi-lingual speeches and stylistic characters. Best: \textbf{bolded}, and second: \underline{underlined}.}
    \vspace{0.4cm}
    \label{tab:newbench}
    \resizebox{\linewidth}{!}{%
    \setlength{\tabcolsep}{3pt}
    \begin{tabular}{@{}l|ccccc@{}}
      \toprule
      Methods & FID $\downarrow$ & FVD $\downarrow$ & HyperIQA $\uparrow$ & LSE-C $\uparrow$ & AVS $\uparrow$ \\
      \midrule
      MuseTalk \cite{zhang2024musetalk}   & 28.41 & 312.57 & 62.34 & 4.66 & 5.87 \\
      Diff2Lip \cite{mukhopadhyay2024diff2lip}   & 33.10 & 358.92 & 58.76 & 7.01 & 5.21 \\
      LatentSync \cite{li2024latentsync}  & 26.32 & 287.43 & 64.51 & 5.87 & 6.34 \\
      OmniSync \cite{peng2025omnisync}    & \underline{24.08} & \underline{263.15} & \underline{67.23} & \underline{3.01} & \underline{6.82} \\
      \textbf{Ours}                       & \textbf{18.34} & \textbf{198.71} & \textbf{72.86} & \textbf{7.14} & \textbf{7.95} \\
      \bottomrule
    \end{tabular}%
    }
\end{minipage}
\vspace{-0.5cm}
\end{table}

%% file: figs/07_qual_samples_and_multi-lingual.tex
\begin{figure}[t]
\centering
\begin{minipage}[t]{0.48\textwidth}
    \centering
    \includegraphics[width=\linewidth]{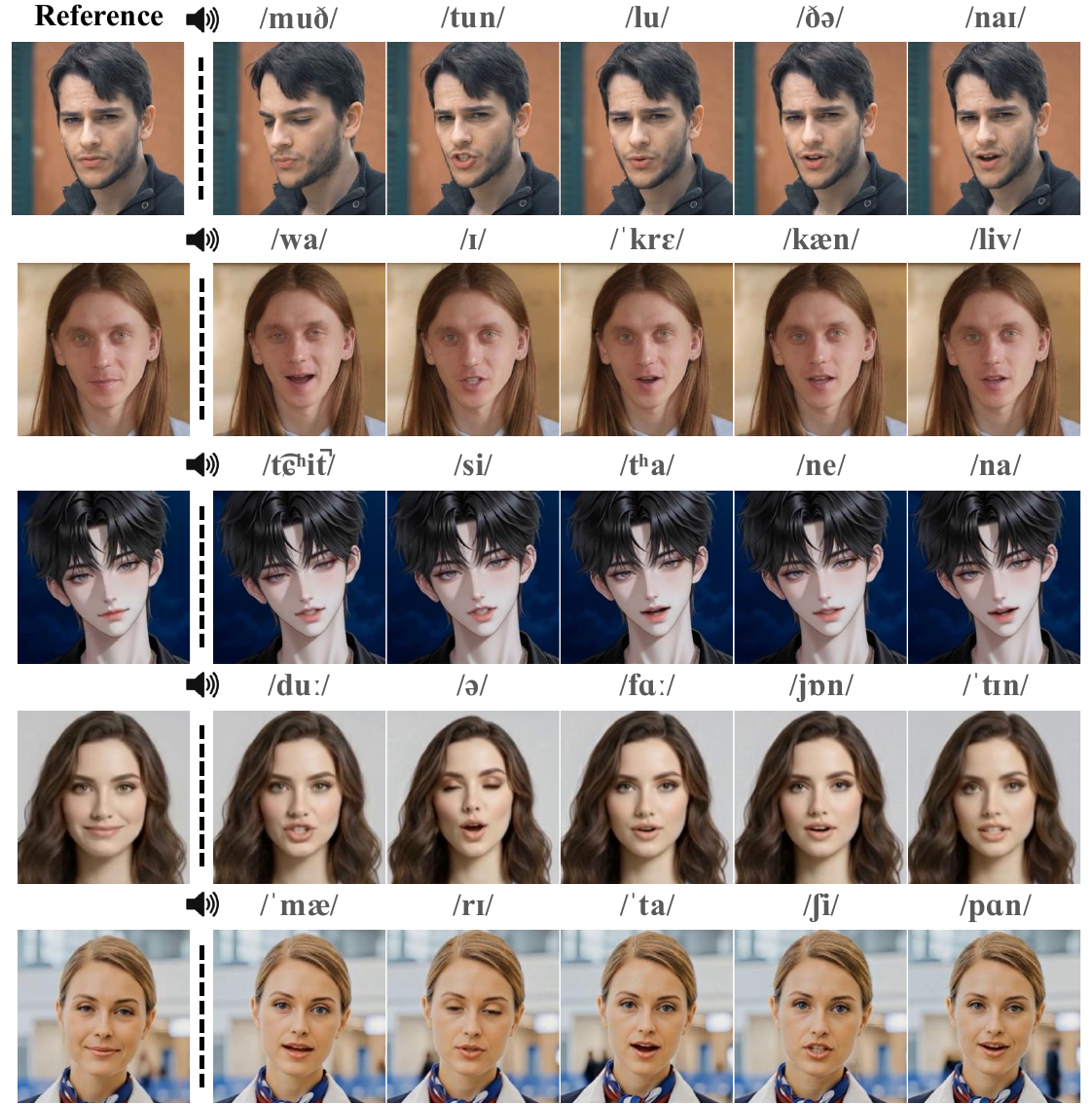}
    \vspace{0.2cm}
    \caption{Qualitative results of our method. It faithfully generates lip-synchronized videos according to driving audios across subjects.}
    \label{fig:qual_samples}
\end{minipage}%
\hfill
\begin{minipage}[t]{0.48\textwidth}
    \centering
    \includegraphics[width=\linewidth]{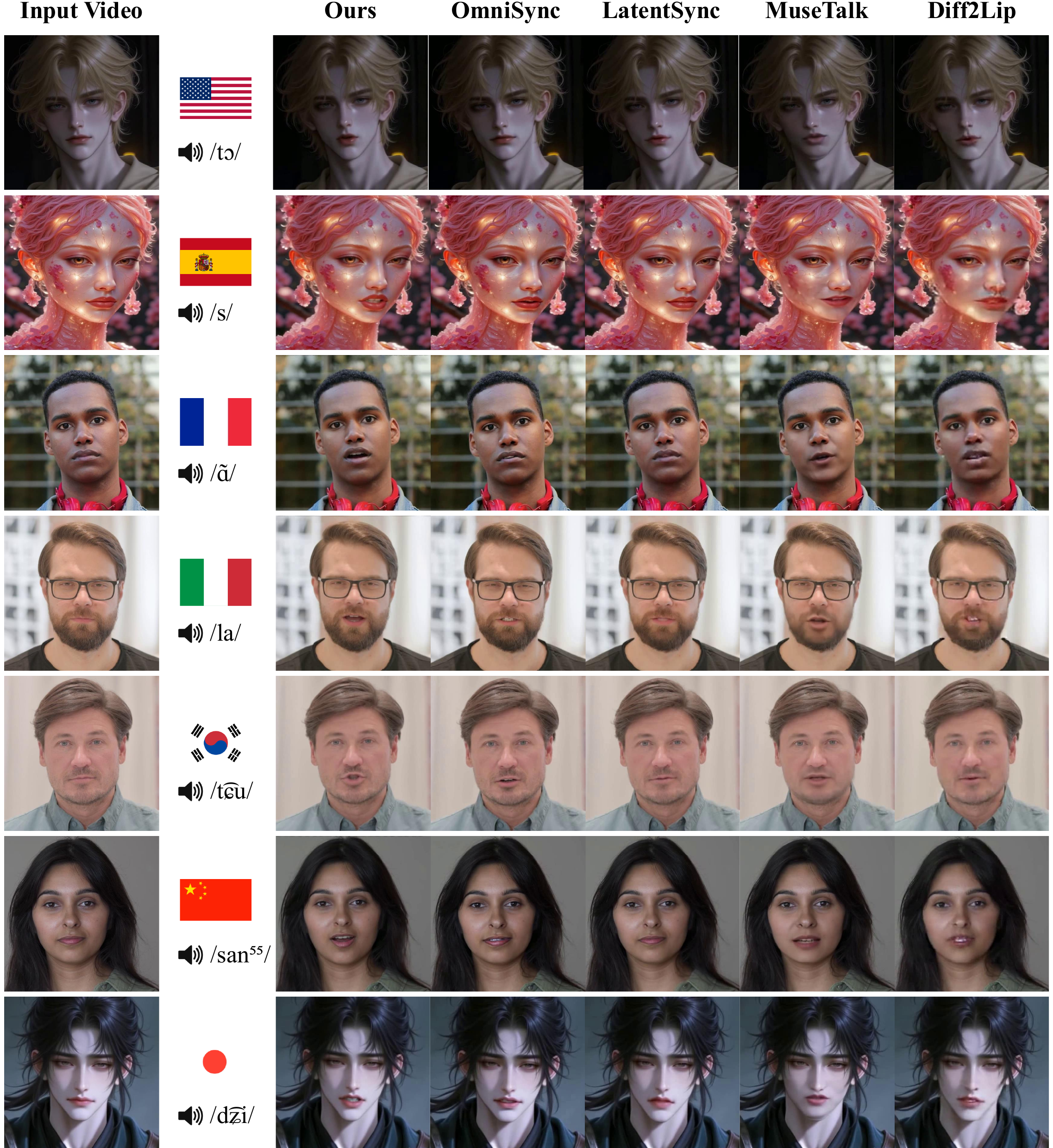}
    \vspace{0.2cm}
    \caption{Qualitative comparison on  UniVoxBench. Our method exhibit robust lip synchronization on multi lingual speeches across human/stylistic characters.}
    \label{fig:multi-lingual1}
\end{minipage}
\vspace{-0.6cm}
\end{figure}

%% file: figs/08_qual_expression_and_userstudy.tex
\begin{figure}[t]
\centering
\begin{minipage}[c]{0.54\textwidth}
    \centering
    \includegraphics[width=\linewidth]{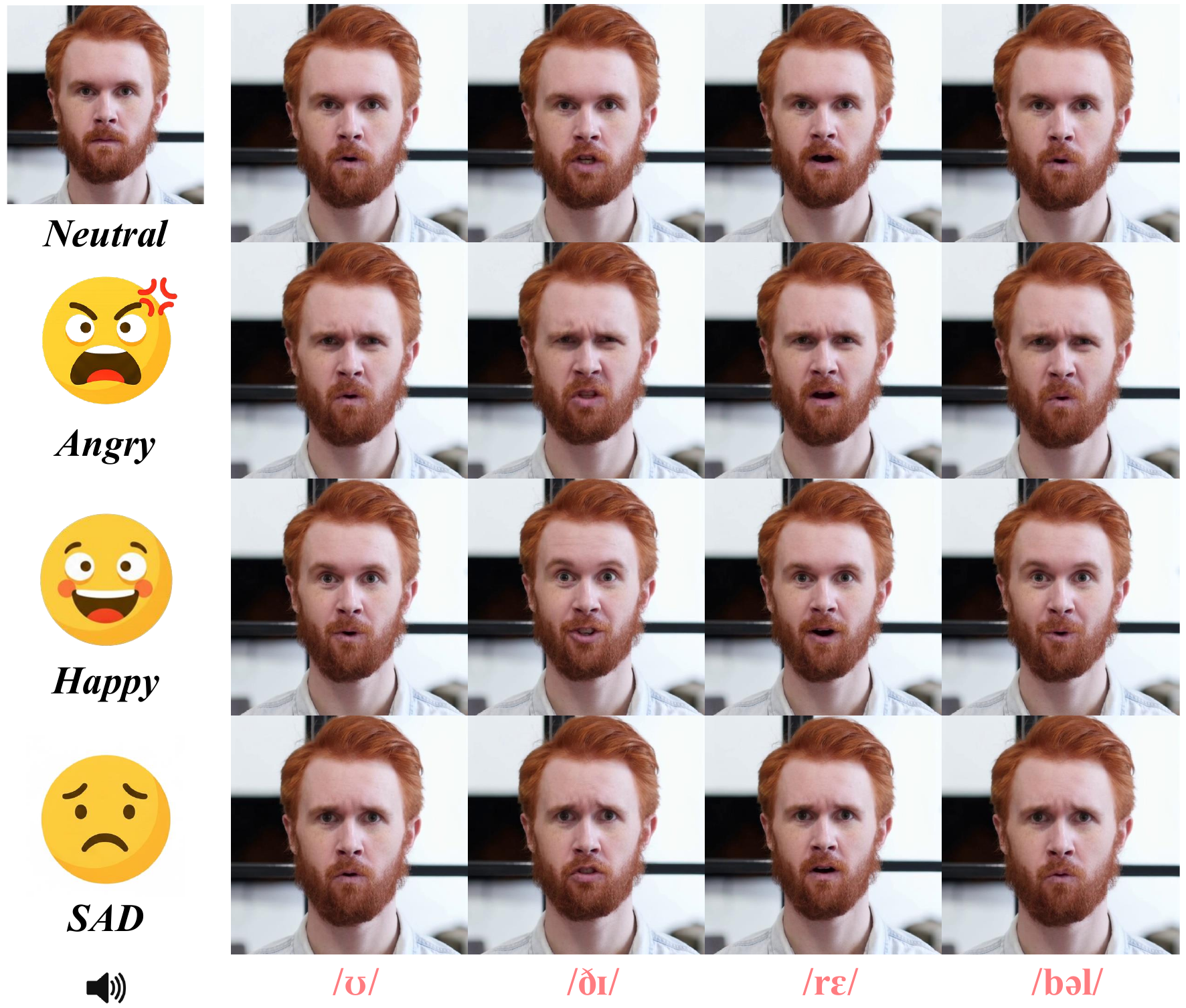}
    \vspace{0.4cm}
    \caption{The effectiveness of expressive control gate. It enables explicit emotion control for diverse and enriched content generation.}
    \label{fig:expression}
\end{minipage}
\hfill
\begin{minipage}[c]{0.42\textwidth}
    \centering
    \includegraphics[width=\linewidth]{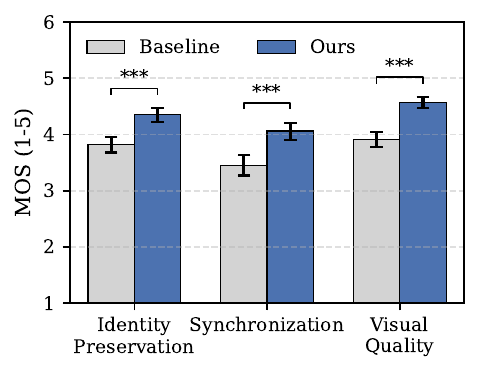}
    \vspace{0.4cm}
    \caption{User study comparing our method with the state-of-the-art baseline across identity preservation, synchronization, and visual synthesis. Our method consistently outperforms the baseline in all metrics; Error bar: std. err., ***: $p < 0.001$.}
    \label{fig:user_study}
\end{minipage}
\vspace{-0.3cm}
\end{figure}

%% file: figs/09_abl_loss_and_arch.tex
\begin{figure}[t]
\centering
\begin{minipage}[t]{0.48\textwidth}
    \centering
    \includegraphics[width=\linewidth]{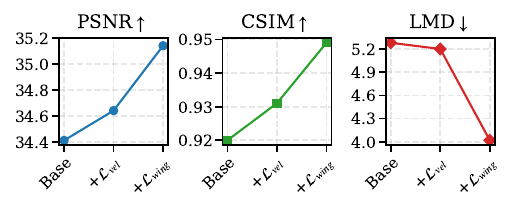}
    \vspace{-0.3cm}
    \caption{
    Ablation study of each loss component $\mathcal{L}_{vel}$ and $\mathcal{L}_{wing}$ for audio-to-implicit motion generation. Both terms prove essential for knowledge distillation of the audio-motion mapping, with removal of either causing notable sync degradation.
    }
    \label{fig:abl_loss}
\end{minipage}%
\hfill
\begin{minipage}[t]{0.48\textwidth}
    \centering
    \includegraphics[width=\linewidth]{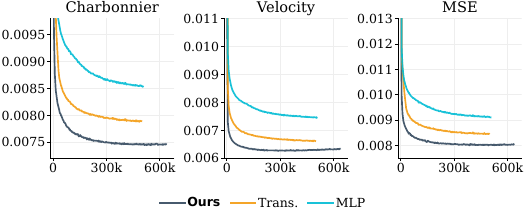}
    \vspace{-0.3cm}
    \caption{
    Ablation study of architecture design choices for the audio-to-implicit motion generation model. The proposed architecture demonstrates superior convergence stability and optimality over alternative design configurations.
    }
    \label{fig:abl_arch}
\end{minipage}
\vspace{-0.4cm}
\end{figure}

%% file: figs/abl_ecg.tex
\begin{figure}[t]
  \centering
  \includegraphics[width=0.99\linewidth]{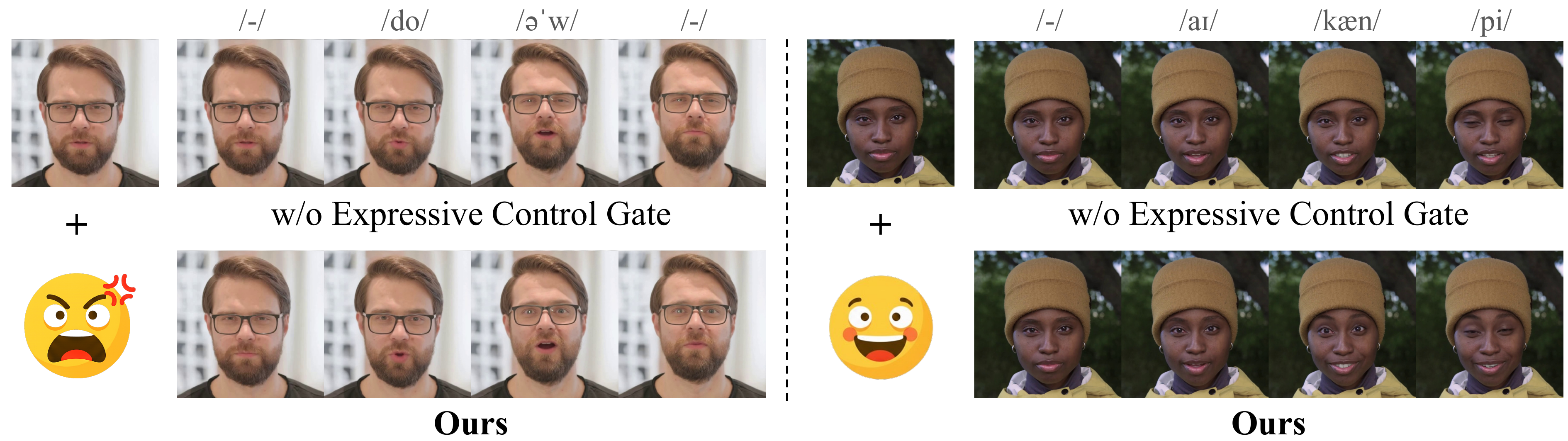}
  \vspace{0.3cm}
  \caption{The visual effectiveness on the proposed expressive control gate.}
  \label{fig:abl_ecg}
  \vspace{-0.5cm}
\end{figure}

%% file: figs/userstudy_emotion.tex
\begin{wrapfigure}[15]{r}[0pt]{0.5\textwidth}
    \centering
    \includegraphics[width=\linewidth]{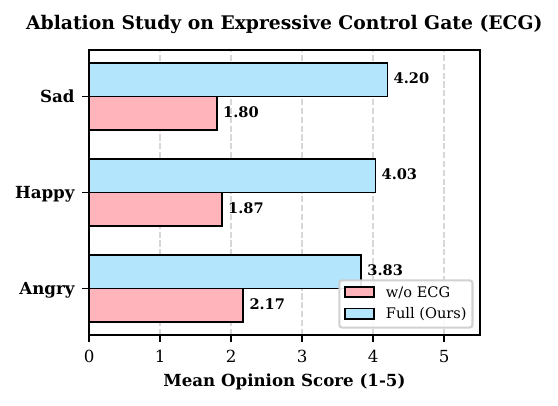}
    \caption{Perceptual ablation on the expressive control gate.}
    \label{fig:emotion_userstudy}
\end{wrapfigure}

%% file: figs/10_qual_comparisons_sub.tex
\begin{figure*}[t!]
    \centering
    \includegraphics[width=0.99\linewidth]{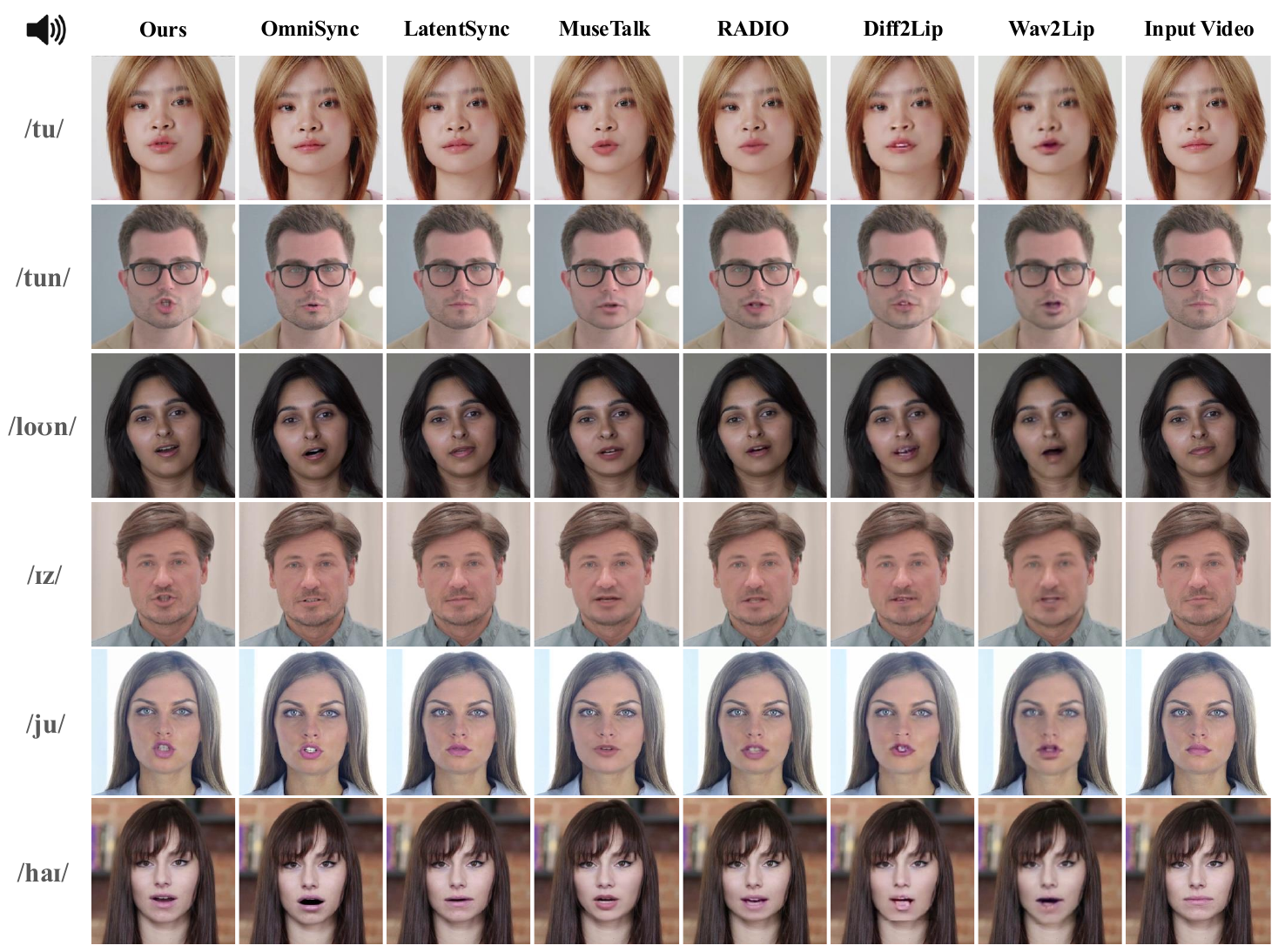}
    \vspace{0.2cm}
    \captionof{figure}{Qualitative comparison with previous lip-sync video methods on human subjects.}
    \label{fig:comparison_human}
    \includegraphics[width=0.99\linewidth]{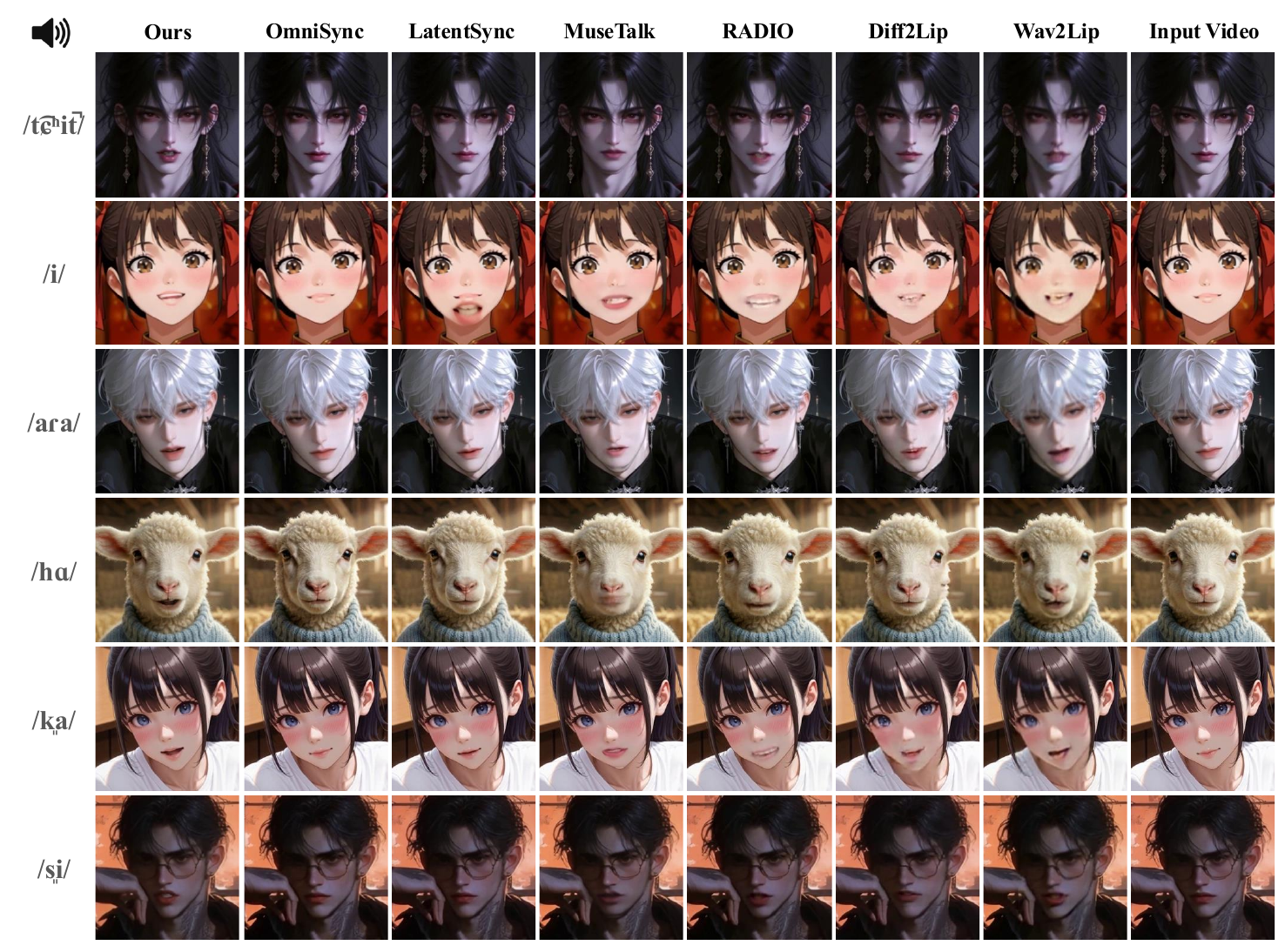}
    \vspace{0.2cm}
    \captionof{figure}{Qualitative comparison with previous lip-sync video methods on characters.}
    \label{fig:comparison_char}
\end{figure*}